# An interpretable deep learning method for early hepatocellular carcinoma diagnosis on multiphase CT

Emma Sarfati, PhD [1,2]

Alexandre Bône, PhD [1,*]

Pietro Gori, PhD [2]

Sisi Yang, MD [1,3,4]

Marc-Michel Rohé, PhD [1]

François Nicolas, PhD [1]

Jeong-Min Lee, MD [5]

Jeong Hee Yoon, MD [5]

Maxime Ronot, MD, PhD [6,7]

Isabelle Bloch, PhD [2,8]

Christophe Aubé, MD, PhD [9,10]

[1] Guerbet Research, Villepinte, France

[2] LTCI, Télécom Paris, Institut Polytechnique de Paris, Saclay, France

[3] MAP5 (UMR 8145), Université Paris Cité, 45 rue des Saints-Pères, Paris, France

[4] Radiology B Department, Hôpital Cochin, 27 Rue du Faubourg Saint-Jacques, Paris, France

[5] Seoul National University Hospital, Department of Radiology, Seoul, Korea

[6] Service de Radiologie, Hôpital Beaujon, APHP-Nord, Clichy, France

[7] Université Paris Cité, CRI, INSERM UMR 1149, Paris, France

[8] Sorbonne Université, CNRS, LIP6, Paris, France

[9] Département de Radiologie CHU Angers, Angers 49933, France

[10] Laboratoire HIFIH, Université d'Angers, Angers, France

[*] Corresponding author. alexandre.bone@guerbet.com, +33 6 12 57 03 64.

Guerbet Research, 15 Rue des Vanesses, 93420 Villepinte, France.

# ABSTRACT

Early diagnosis of hepatocellular carcinoma is critical to preserve access to curative treatment options. The Liver Imaging Reporting and Data System (LI-RADS) provides a standardized framework for non-invasive diagnosis on multiphase computed tomography, based on major visual imaging features. However, these features may be difficult to assess consistently, particularly in small liver lesions. We developed dLI-RADS, an interpretable artificial intelligence method designed to estimate the probability of hepatocellular carcinoma from multiphase computed tomography images of manually segmented liver lesions, while remaining aligned with LI-RADS-related imaging concepts. dLI-RADS combines neural network-derived predictions with handcrafted quantitative imaging measurements reflecting LI-RADS major features and integrates these inputs using logistic regression. In this multicenter retrospective study, patients with multiphase computed tomography and at least one liver lesion were included, while lesions larger than 40 mm were excluded. Histology served as the reference standard for hepatocellular carcinoma, and a composite reference standard was used for non-hepatocellular carcinoma lesions. The lesion-level accuracy of dLI-RADS and LI-RADS were compared on the internal evaluation set using 10,000-repetition bootstrap non-inferiority tests, with 5% margins and significance levels. On the external test set, sensitivities were similarly compared. A total of 143 (61 years ±9, 125 men), 62 (61y±10, 49 men), and 133 (59y±10, 103 men) patients with 196, 84, and 171 lesions (mean diameter 17.7, 18.3, and 21.8 millimeters) were respectively included in the training, internal and external evaluation sets. dLI-RADS was non-inferior to LI-RADS for lesion-level accuracy in the internal evaluation set, achieving 80% accuracy compared with 74% for LI-RADS ($p = 0.04$). On the external evaluation set, dLI-RADS achieved a sensitivity of 93% compared with 78% for LI-RADS (non-inferiority $p < 0.001$). On this same set, dLI-RADS also achieved 74% specificity, 86% accuracy, and an AUC of 84%. Because LI-RADS features were unavailable for non-HCC lesions, only sensitivity could be directly compared between dLI-RADS and LI-RADS. These findings suggest that dLI-RADS may provide interpretable decision support for hepatocellular carcinoma diagnosis on multiphase computed tomography, particularly for small liver lesions.

# Introduction

Accurate and timely diagnosis of hepatocellular carcinoma (HCC) is essential to preserve treatment options [1]. Since its original introduction to standardize the interpretation of multi-phase imaging and provide a common probabilistic language for clinical decision-making, the Liver Imaging Reporting and Data System (LI-RADS) v2018 [2] has become a reference methodology for HCC diagnosis [3,4], with evidence of high accuracy levels [5]. Nonetheless, non-invasive HCC diagnosis remains challenging and subject to inter-reader variability, particularly on CT where major features are more difficult to assess than on MRI [6–8].

In parallel to these efforts to standardize the usage of long-known imaging characteristics of HCC, novel quantitative markers were proposed in “radiomics” studies [9,10] where pre-determined libraries of handcrafted features are mined to identify promising diagnostic or prognostic elements, with the intent to automate, complement or even surpass human image analysis. Going a step further in the terms of abstraction, deep learning methods were also proposed to extract purely data-driven HCC features, reporting promising diagnostic performance levels [11–20]. However, a common fundamental limitation of these studies is their focus on large-size HCC lesions, i.e. with average diameters larger than 30 mm, while there is a strong interest in diagnosing smaller early-stage HCC nodules, which are much more difficult to correctly characterize [8,21,22]. These approaches are also limited in their explainability: HCC presence probabilities are typically computed without any details about the reasoning process, or at best with spatial activation maps [14] which allow the user to visually verify that the correct region of the image (i.e. the tumor) contributed to the prediction. This contrasts with handcrafted radiomics that are intrinsically interpretable.

In this study, we propose and evaluate dLI-RADS, a deep learning method for lesion-level diagnosis of early-stage HCC from manually segmented multiphase CT lesions. Methodologically, dLI-RADS relies on a combination of deep and handcrafted features to strike a balance between clinical performance and model explainability. This method was specifically designed to perform HCC diagnosis for challenging small-size liver lesions, and its performance was compared to LI-RADS using non-inferiority tests in two independent cohorts of patients from Europe and Asia respectively.

# Methods

## Study Design and Datasets

Two cohorts were leveraged in this retrospective multicentric study. The main cohort includes 442 patients with compensated cirrhosis enrolled at 16 European centers between 2009 and 2012. All *de novo* liver nodules underwent contrast-enhanced CT and histological assessment; a composite reference standard resolved non-malignant cases. Two senior radiologists independently assessed LI-RADS v2018 major features [2]; two additional radiologists performed 3D portal-phase lesion segmentation.

The secondary cohort includes 513 patients with de novo liver nodules (2008–2018, large Asian center). The reference standard was histology for malignant tumors and ≥6-month follow-up for benign lesions. Radiologists (≥3 years of experience) performed portal-phase 3D segmentations and assessed LI-RADS major features. In both cohorts, exams lacking arterial, portal, or delayed phase were excluded, as well as lesions without definitive diagnosis, larger than 40 mm, or classified as LR-1, LR-2, or LR-M. LI-RADS classification, including LR-M assessment, was performed by radiologists blinded to the reference diagnosis and according to LI-RADS v2018 guidelines. LR-M exclusion was based on radiological classification rather than pathological diagnosis; consequently, pathologically confirmed metastases or cholangiocarcinomas remained eligible when they had not been classified as LR-M on imaging.

The main cohort was split 70/30 into training and internal evaluation sets with strict patient-level segregation. The secondary cohort, with no patient overlap, served as an external evaluation set.

### Ethics declarations

For the main cohort, the study was registered as #NCT00848952 and previously published in [23]. Informed consent was obtained from all participants. For the secondary cohort, the local institutional review board of Seoul National University Hospital approved the retrospective study and waived the requirement for informed consent. All methods were performed in accordance with relevant guidelines and regulations and with the principles of the Declaration of Helsinki. All data were de-identified before analysis.

## The dLI-RADS method

### *Overview*

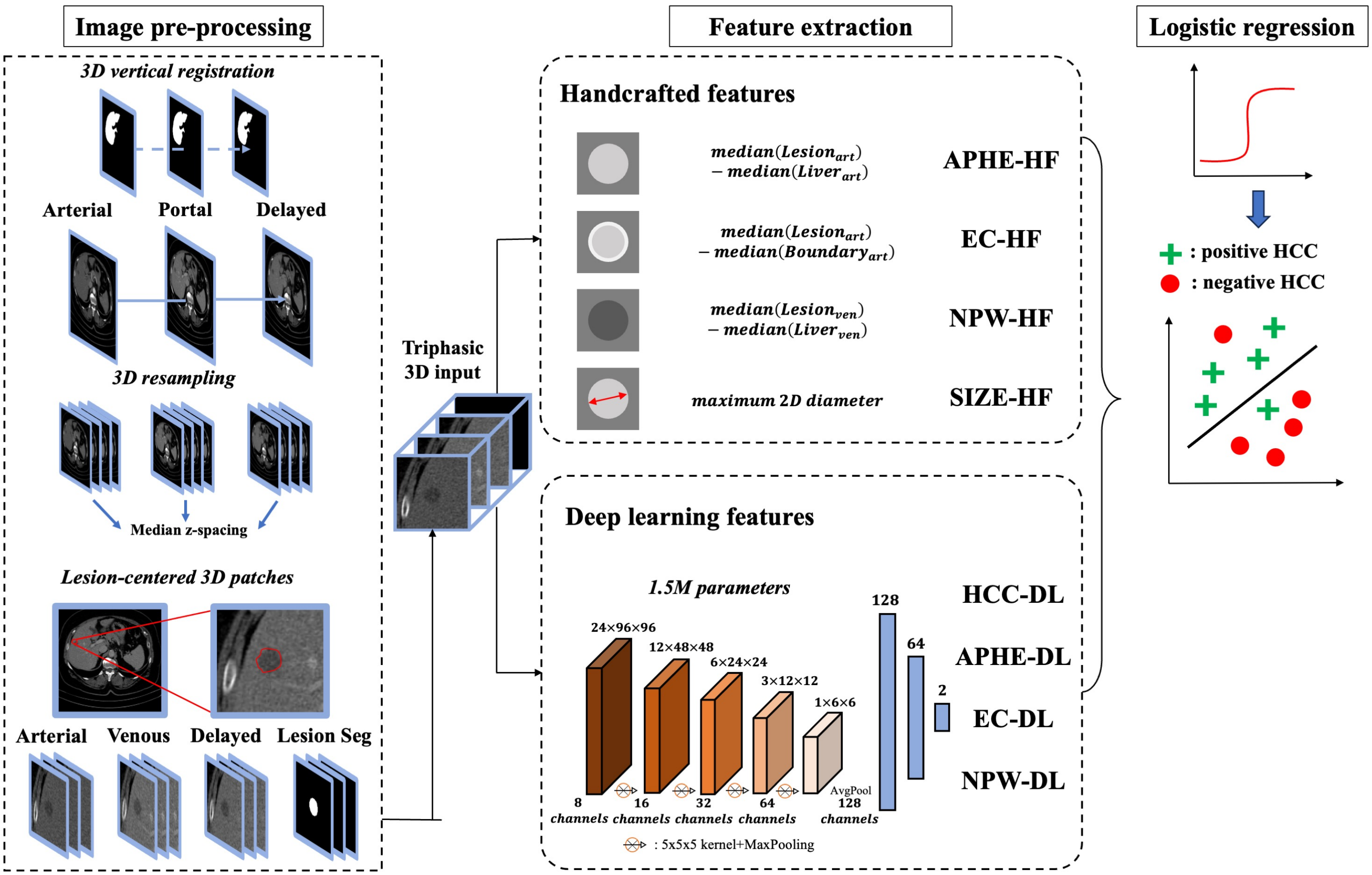

**Figure 1**. Overview of the dLI-RADS method. Tri-phasic computed tomography patches and lesion masks are used to compute handcrafted descriptors and deep learning-derived probabilities related to LI-RADS major features and HCC presence. These intermediate features are combined with logistic regression to predict lesion-level HCC risk.

Fig.1 details the proposed dLI-RADS method, which processes tri-phasic CT image patches centered around pre-segmented liver lesion candidates to predict a probability of HCC presence as a raw output.

In a first step, a series of 8 intermediate features are extracted with two families of approaches. Three handcrafted (HF) features are computed using predefined formulas, that strive to respectively encode the arterial phase hyperenhancement (APHE), enhancing capsule (EC), and non-peripheral washout (NPW) major LI-RADS features: APHE-HF is defined as the difference in median arterial-phase Hounsfield values between the lesion and the liver parenchyma, EC-HF as the difference in arterial-phase median signals between the lesion and its 1-voxel-wide inner boundary, and NPW-HF as the difference in venous-phase median signals between the lesion and the liver parenchyma. A fourth HF is the lesion size, directly computed from the lesion segmentation mask. Four deep learning (DL) features correspond to the probabilities predicted by four neural networks, each trained on a specific binary classification task: APHE, EC, NPW, and HCC presence. In a second step, a logistic regression model combines these intermediate features to predict the final probability of HCC presence.

### *Image pre-processing*

CT scans were clipped between -100 and 400 Hounsfield units. Arterial, portal venous, and delayed phase were spatially co-registered with vertical translations to compensate for breathing displacements, using a grid search method that evaluates registration quality as the overlap between fixed and moving liver segmentation masks. All volumes were then resampled to the 2.00 × 0.76 × 0.76 $mm^3$ voxel size. Image patches were extracted for candidate liver lesions, centered around the corresponding segmentation masks. A common patch size of 24 × 96 × 96 voxels was chosen. Arterial, portal and delayed phases as well as the lesion segmentation were stacked in the channel dimension.

### *Training and inference*

The four neural networks shared the same architecture that is detailed in Fig.1: five convolutional blocks with 5 × 5 × 5 kernels and 2 × 2 × 2 max pooling, followed by two dense layers and a softmax output. Rectified linear unit activations were used after each convolutional or dense layer. This architecture was directly inspired by [12], in which the authors reported an AUC of 92% for HCC classification. The networks were trained for 300 epochs with a batch size of 32 using the AdamW optimizer. The learning rate was $10^{-5}$ and the weight decay was $10^{-3}$. Training was performed on a Tesla V100 graphics processing unit. Deep learning models were trained using PyTorch-Lightning (v2.5), and the logistic regression was calibrated using scikit-learn (v1.7). Handcrafted features were computed using scikit-image (v0.25) and pyradiomics (v3.1).

After training, the neural networks were applied to the training images to derive deep learning features. These features were concatenated with the handcrafted descriptors to fit the logistic regression model. The raw logistic regression output was a continuous HCC probability between 0 and 1. To define clinically interpretable operating points, thresholds were chosen to minimize the sum of the absolute distance to radiologists' sensitivity and specificity on the training set, for both LI-RADS-5 and LI-RADS-4,5 points:

$$dLR - k = \underset{p \in [0,1]}{\operatorname{argmin}} |SE(p) - SE_{LR-k}| + |SP(p) - SP_{LR-k}|$$

This procedure gave us two thresholds, that we analogically named dLI-RADS-4,5 and dLI-RADS-5, directly comparable to LI-RADS-4,5 and LI-RADS-5. This strategy is related to index-of-union threshold selection for receiver operating characteristic curves [26] but was adapted here to match established clinical operating points.

## Statistical analysis

The diagnostic performance of dLI-RADS and LI-RADS were evaluated using sensitivity (SE), specificity (SP), and accuracy (ACC). These metrics were derived for the dLR-5 and LR5 groups, as well as for the unions of the dLR-4 and dLR-5 (noted dLR-4,5), LR-4 and LR-5 (noted LR-4,5) groups respectively. Only the sensitivity of LI-RADS could be evaluated on the external evaluation, as major features were not collected for non-HCC lesions. The AUC performance of the raw dLI-RADS predicted probability was also reported. Bootstrap 95% confidence intervals were built for all metrics using 10,000 repetitions. Non-inferiority of dLI-RADS vs. LI-RADS was tested using non-parametric bootstrap *p*-values [24,25]. An absolute non-inferiority margin of 5 percentage points was selected consistently with previous deep learning studies in closely related HCC diagnostic settings using a 5% margin for comparisons with conventional or radiological assessment [14,27]. This margin was considered an appropriately stringent tolerance to limit the acceptable decrease in diagnostic performance relative to LI-RADS. A significance level of 5% was used. All reported comparative *p*-values correspond to non-inferiority testing and should not be interpreted as tests of statistical superiority.

To assess complementarity between dLI-RADS and LI-RADS, compound risk groups were defined by summing their corresponding risk categories, from (dLR+LR)-6 to (dLR+LR)-10. Feature importance was explored using ablation experiments. For each number of retained intermediate features, all possible combinations were evaluated during model selection on the training set, and the best-performing combination was evaluated on the internal evaluation set.

# Results

## Patient and lesion characteristics

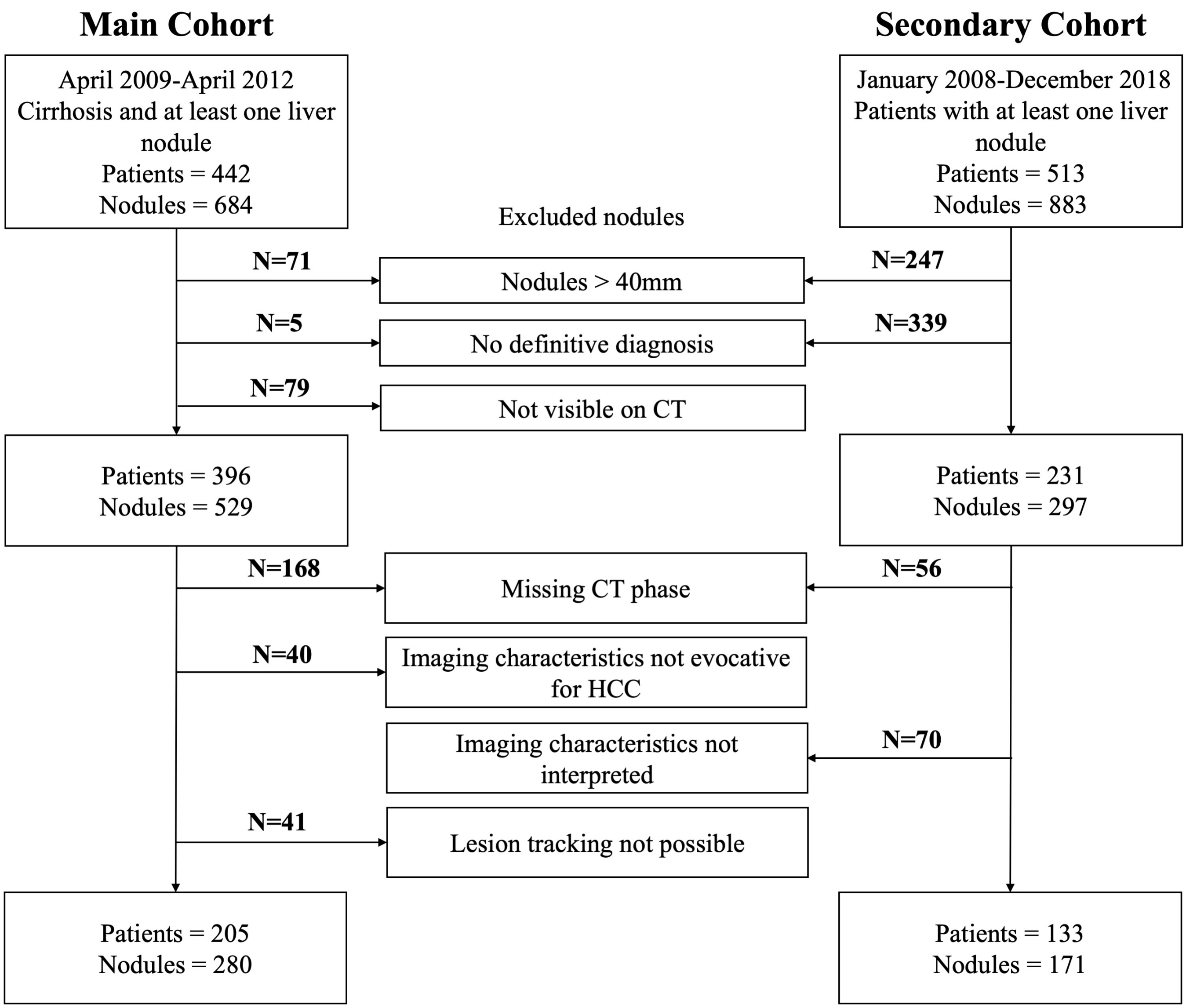

**Figure 2**. Flowchart of patient inclusion. The diagram shows the inclusion and exclusion steps for the main and external cohorts, including exclusions for incomplete multiphase computed tomography, lesion size greater than 40 mm, missing definitive diagnosis, and LI-RADS inapplicability.

| | Training set (*N=143*) | Internal evaluation set (*N=62*) | External evaluation set (*N=133*) |
|---|---|---|---|
| Age, years, range | 61.2 (40.2,82.2) | 61.3 (41.2,81.1) | 58.9 (32.1,88.5) |
| Sex (Male / Female / Unknown) | 87.4% / 12.6% / 0% (125 / 18 / 0) | 79.0% / 21.0% / 0.0% (49 / 13 / 0) | 77.4% / 22.6% / 0.0% (103 / 30 / 0) |
| Number of centers | 16 | 16 | 1 |
| Liver cirrhosis (Yes / No / Unknown) | 100.0% / 0.0% / 0.0% (143 / 0 / 0) | 100.0% / 0.0% / 0.0% (62 / 0 / 0) | 88.7% / 3.8% / 7.5% (118 / 5 / 10) |
| Risk factor | | | |
| Alcohol (Yes / No / Unknown) | 57.3% / 41.3% / 1.4% (82 / 59 / 2) | 42.0% / 58.0% / 0% (26 / 36 / 0) | 8.3% / 91.7% / 0.0% (11 / 122 / 0) |
| Metabolic dysfunction-associated steatohepatitis (Yes / No / Unknown) | 18.9% / 79.7% / 1.4% (27 / 114 / 2) | 13.0% / 87.0% / 0.0% (8 / 54 / 0) | 67.7% / 21.8% / 10.5% (90 / 29 / 14) |
| Hepatitis B (Yes / No / Unknown) | 5.6% / 93% / 1.4% (8 / 133 / 2) | 6.5% / 93.5% / 0.0% (4 / 58 / 0) | 76.7% / 13.5% / 9.8% (102 / 18 / 13) |
| Hepatitis C (Yes / No / Unknown) | 21% / 77.6% / 1.4% (30 / 111 / 2) | 35.5% / 64.5% / 0.0% (22 / 40 / 0) | 8.3% / 82% / 9.7% (11 / 109 / 13) |
| Multi-factor (Yes / No / Unknown) | 10.5% / 88.1% / 1.4% (15 / 126 / 2) | 11.3% / 88.7% / 0.0% (7 / 55 / 0) | Missing values |
| Child-Pugh (A / B / C / Unknown) | 69.9% / 20.9% / 6.1% / 3.1% (137 / 41 / 12 / 6) | 69% / 19% / 7.1% / 4.9% (58 / 16 / 6 / 4) | Missing values |
| Lesions per patient (Mean ± Std) | 1.4 ± 0.6 | 1.4 ± 0.6 | 1.3 ± 1.0 |
| Lesion size (Mean ± Std) (mm) | 17.7 ± 6.0 | 18.3 ± 5.8 | 21.8 ± 9.5 |
| Lesion volume (Mean ± Std) ($cm^3$) | 4.5 ± 4.0 | 6.1 ± 6.0 | 4.9 ± 5.1 |
| Lesion count (N / %)<br><br>Lesion size (Mean ± Std) | | | |
| Hepatocellular carcinoma (HCC) | 130 / 66.3% 18.8 ± 5.8 | 58 / 69.0% 19.7 ± 5.5 | 106 / 62.0% 26.0 ± 6.8 |
| Simple cyst or lipoma | 7 / 3.6% 14.1 ± 3.6 | 4 / 4.8% 18.5 ± 7.4 | 39 / 22.8% 9.7 ± 2.7 |
| Metastasis | 14 / 7.1% 18.6 ± 8.4 | 5 / 6% 13.2 ± 2.1 | 5 / 2.9% 18.4 ± 8.4 |
| Hemangioma | 30 / 15.3% 14.8 ± 5.1 | 14 / 16.7% 15.1 ± 5.5 | 5 / 2.9% 15.9 ± 6.2 |
| Cholangiocarcinoma | / | / | 14 / 8.2% 27.1 ± 10.8 |
| Focal nodular hyperplasia | 1 / 0.5% 21.0 ± 0.0 | / | 1 / 0.6% 20.3 ± 0.0 |
| Perfusion defect | / | / | 1 / 0.6% 12.1 ± 0.0 |
| Other | 14 / 7.1% 13.7 ± 3.9 | 3 / 3.6% 16.5 ± 5.0 | / |

**Table 1**. Clinical characteristics of patients in training, internal evaluation, and external evaluation sets.

Fig.2 details the inclusion results. 205 patients were finally included in the main cohort, from 442 initially recruited. The secondary cohort consists of 133 patients from the 513 initially recruited.

A total of 338 patients were ultimately included. Table 1 details their characteristics. The training, internal evaluation and external evaluation sets were respectively composed of 143 (mean age 61 years ± 9, 125 men), 62 (mean age 61 years ± 10, 49 men), and 133 (mean age 59 years ± 10, 103 men) patients. All patients in the training and internal evaluation sets had cirrhosis, and excessive alcohol consumption was the most frequent risk factor for liver disease. In the external evaluation set, 89% of patients had cirrhosis, and the most frequent risk factor for liver disease was hepatitis B infection. In terms of liver lesions, the training, internal evaluation and external evaluation sets were respectively composed of 196 (mean size 18 mm ± 6, 130 HCC), 84 (mean size 18 mm ± 6, 58 HCC) and 171 nodules (mean size 22 mm ± 10, 106 HCC).

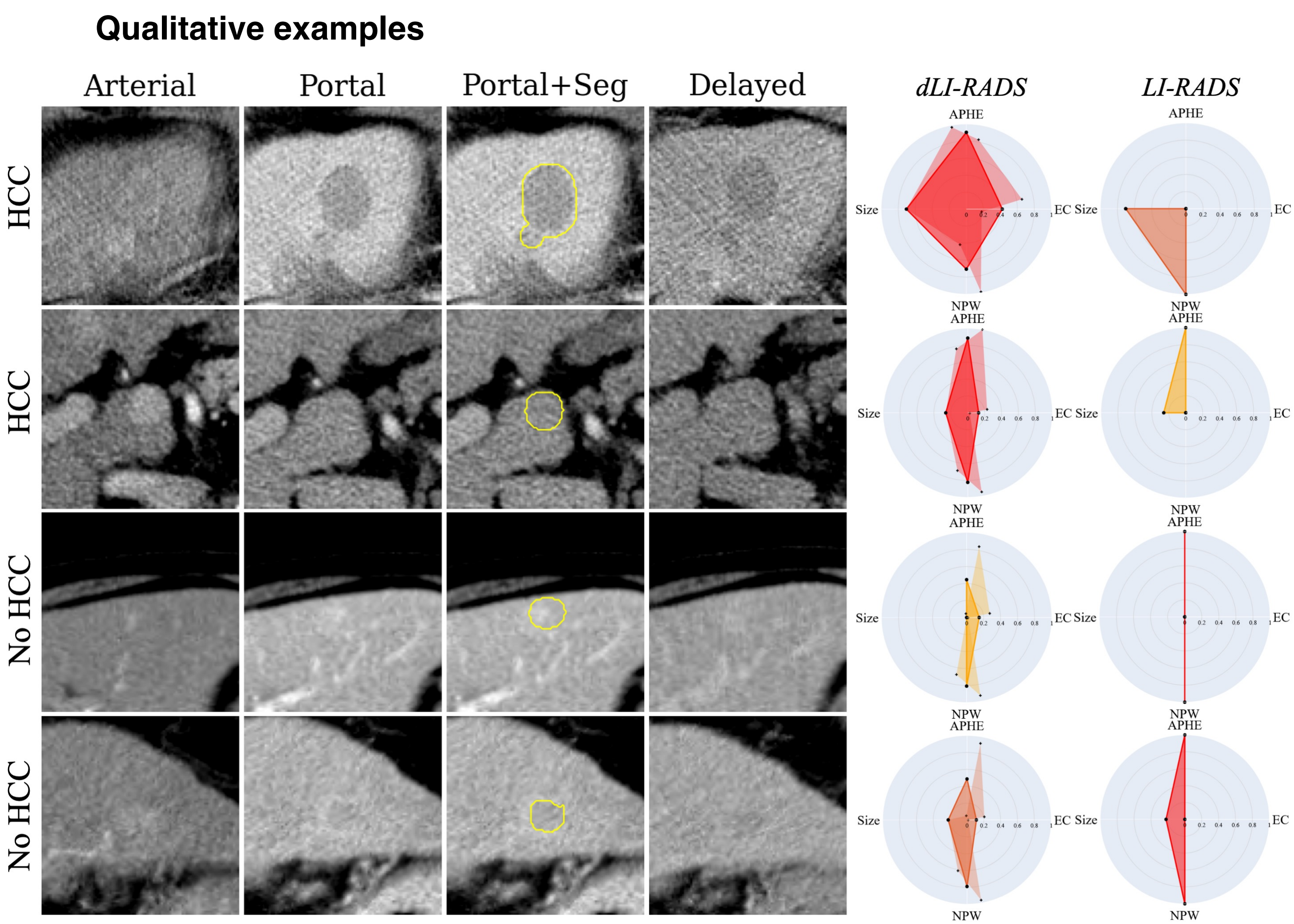

**Figure 3**. Examples of liver lesions analyzed by dLI-RADS and LI-RADS. Representative computed tomography patches are shown with corresponding radar plots summarizing model-

derived and radiological features. The figure illustrates concordant and discordant assessments between dLI-RADS and LI-RADS.

Fig. 3 provides examples of liver lesions analyzed and classified by the dLI-RADS and LI-RADS methods, on the internal evaluation set. The two top rows correspond to HCC lesions, while the two bottom rows correspond to a hemangioma and a metastasis. On the right-hand side, radar plots display the interpretable intermediate features leveraged by the classification systems to categorize the lesions, and their colors correspond to the selected risk group (risk group 3 in yellow, risk group 4 in orange, risk group 5 in red). The first HCC lesion has a visible washout, that was captured by both dLI-RADS and LI-RADS, and a subtle arterial-phase hyper-enhancement that was not judged sufficient by the annotating radiologist to be included in its LI-RADS assessment but was captured by dLI-RADS. On the two central rows, dLI-RADS and LI-RADS selected opposed risk groups (5 vs. 3 and 3 vs. 5) for the assessed lesions. The last row displays a difficult HCC confounder with APHE, NPW and a mild enhancing capsule which dLI-RADS and LI-RADS respectively characterized as dLR-5 and LR-5.

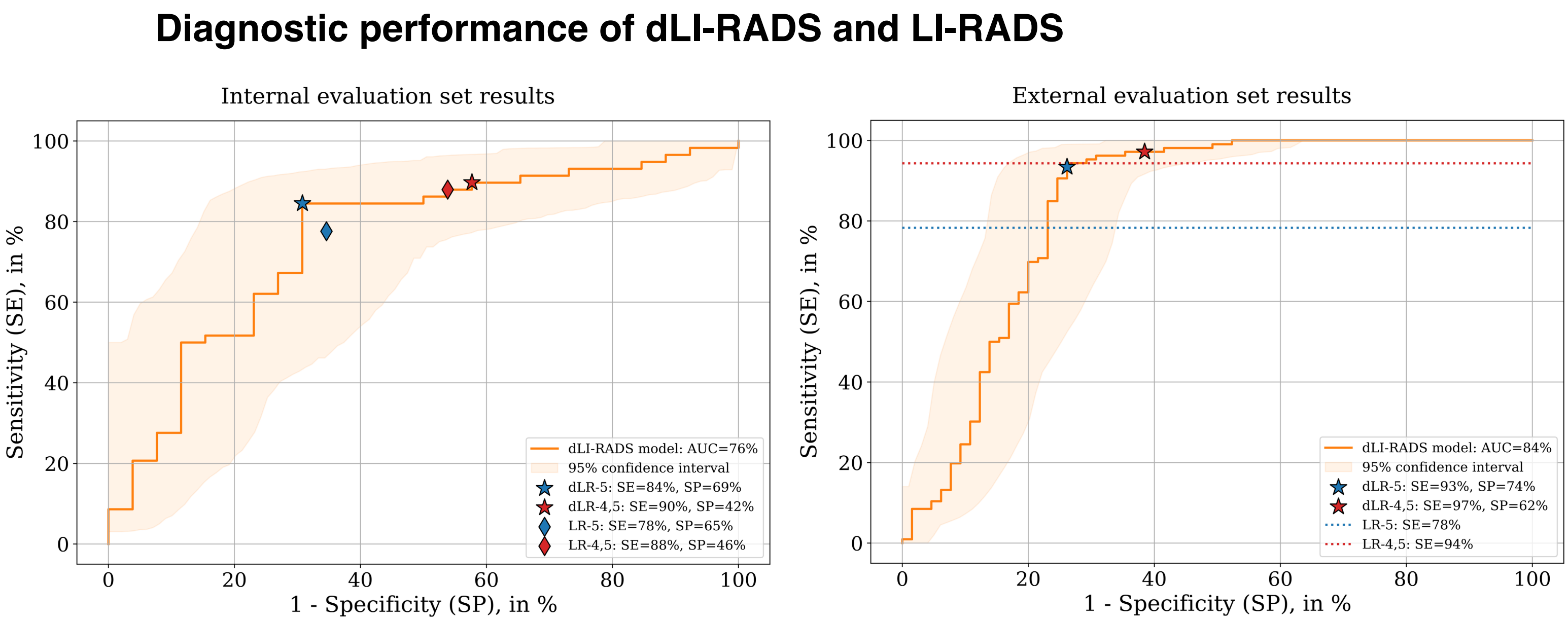

**(a)** **(b)**

**Figure 4**. Receiver operating characteristic curves for dLI-RADS. Curves are shown for the internal **(a)** and external **(b)** evaluation sets. Operating points corresponding to dLR-5 and dLR-4,5 are shown together with the corresponding LI-RADS operating points when available.

| | Internal evaluation set (N=62) | | | | | | External evaluation set (N=133) | | | | | |
|---|---|---|---|---|---|---|---|---|---|---|---|---|
| | dLI-RADS | | LI-RADS | | Non-inferiority p-values | | dLI-RADS | | LI-RADS | | Non-inferiority p-values | |
| | dLR-4,5 | dLR-5 | LR-4,5 | LR-5 | dLR-4,5 vs. LR4,5 | dLR-5 vs LR-5 | dLR-4,5 | dLR-5 | LR-4,5 | LR-5 | dLR-4,5 vs. LR4,5 | dLR-5 vs LR-5 |
| SE | 90 (81,97) [52/58] | 84 (75,93) [49/58] | 88 (79,96) [51/58] | 78 (67,88) [45/58] | 0.10 | **<u>0.03</u>** | 97 (94,100) [103/106] | 93 (88,98) [99/106] | 94 (90,98) [100/106] | 78 (70,86) [83/106] | **<u>0.001</u>** | **<u><0.001</u>** |
| SP | 42 (23,62) [11/26] | 69 (50,86) [18/26] | 46 (27,65) [12/26] | 65 (46,83) [17/26] | 0.44 | 0.26 | 62 (49,73) [40/65] | 74 (63,84) [48/65] | | | | |
| ACC | 75 (66,83) [63/84] | 80 (70,88) [67/84] | 75 (66,83) [63/84] | 74 (64,83) [62/84] | 0.16 | **<u>0.04</u>** | 84 (78,89) [143/171] | 86 (81,91) [147/171] | n/a | | | |
| AUC | 76 (63,87) | | n/a | | | | 84 (76,91) | | | | | |

**<u>Table 2</u>**. Diagnostic performance of dLI-RADS and LI-RADS for HCC diagnosis on internal evaluation and external evaluation sets. Diagnostic performance of LI-RADS (LR) and the proposed dLI-RADS (dLR) are assessed on two risk groups: high-risk (dLR-4,5, LR-4,5) and maximal-risk (dLR-5, LR-5). Performance metrics are sensitivity (SE), specificity (SP), accuracy (ACC), and area under the receiver operating characteristic curve (AUC). All metrics are expressed in percentage points. Between parentheses are given 95% confidence intervals, and between brackets the integer numerator and denominator. Non-inferiority *p*-values of dLI-RADS vs. LI-RADS computed with 10,000-repetition bootstrap testing are also given in dedicated columns. Non-inferiority margin and significance level were both 5%. Bold and underlined *p*-values indicate statistical significance. Reported *p*-values correspond exclusively to non-inferiority testing and should not be interpreted as tests of superiority.

On the internal evaluation set, the maximal-risk groups dLR-5 and LR-5 respectively achieved sensitivities of 84% and 78% ($p = 0.03$ at the 5% non-inferiority margin), specificities of 69% and 65% ($p = 0.26$), and accuracies of 80% and 74% ($p = 0.04$). The high-risk groups dLR-4,5 and LR-4,5 achieved sensitivities of 90% and 88% ($p = 0.10$), specificities of 42% and 46% ($p = 0.44$), and accuracies of 75% ($p = 0.16$). On the external evaluation set, the sensitivities were 93% and 78% fordLR-5 and LR-5 respectively ($p < 0.001$), and 97% and 94% for dLR-4,5 and LR-4,5 ($p = 0.001$). Specificity and accuracy could not be computed for LI-RADS as only HCC nodules were characterized. For dLI-RADS, specificity and accuracy were 74% and 86% for dLR-5, and 62% and 84% for dLR-4,5. These results are reported in Table 2.

Fig.4 plots the receiver operating characteristic curves of the dLI-RADS method on the internal (Fig.4a) and external evaluation (Fig.4b) sets, where the AUC were 76% and 84% respectively. The dLR-5 and dLR-4,5 operating points are represented with star-shaped markers. The diagnostic performance of LR-5 and LR-4,5 is reported with diamond-shaped markers on the internal evaluation set, and dashed lines on the external evaluation set where only sensitivity could be derived.

## Combined model-based and radiological assessment

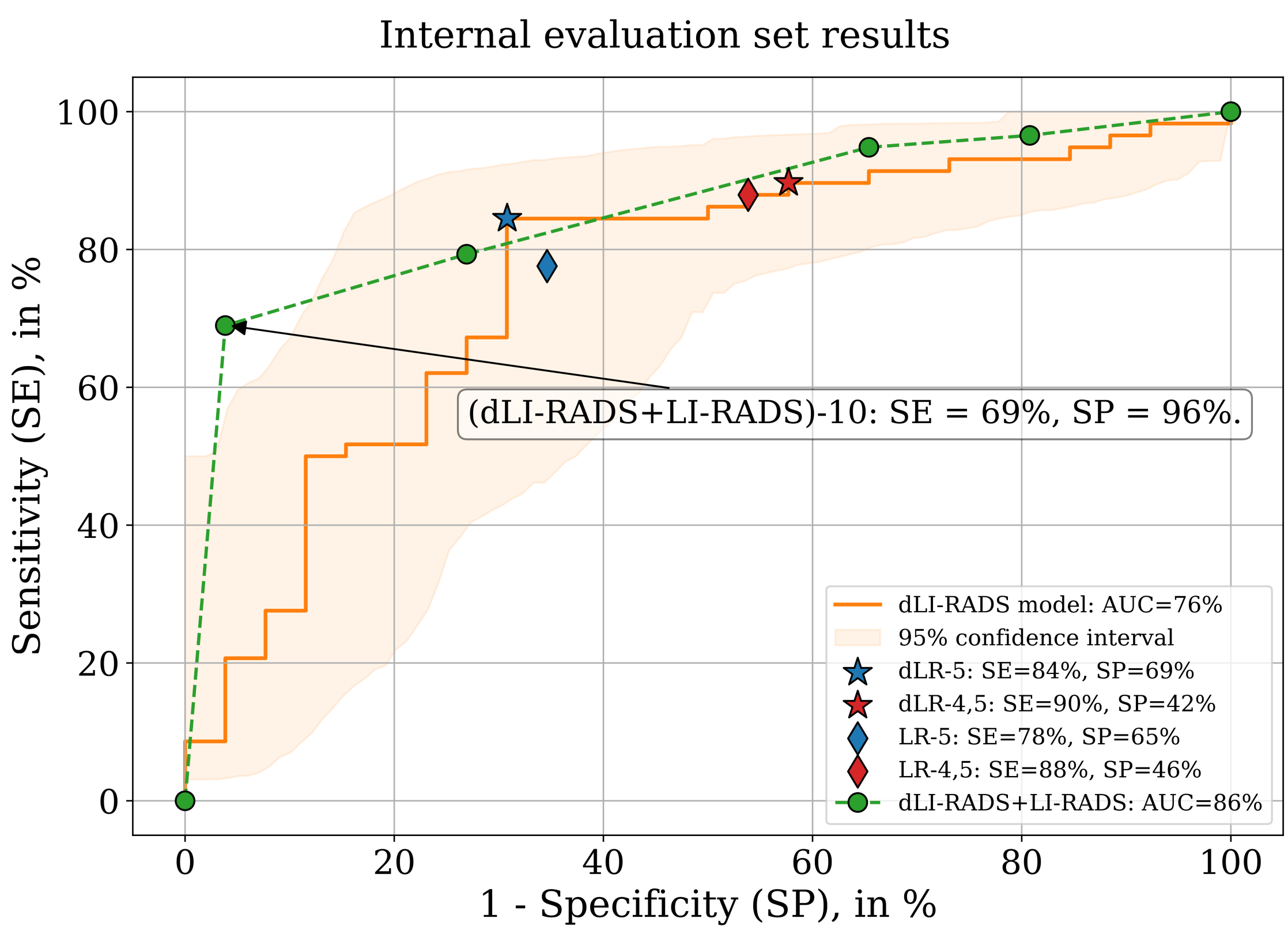

**Figure 5**. Combined dLI-RADS and LI-RADS assessment. Compound risk groups were defined by summing dLI-RADS and LI-RADS categories on the internal evaluation set. The maximal combined category reached high specificity, suggesting possible complementarity between model-based and radiological assessment.

Fig.5 illustrates the diagnostic performance of the combined dLI-RADS and LI-RADS methods on the internal evaluation set. The combined risk groups ranged from (dLR+LR)-6 to (dLR+LR)-10, defining a series of operating points marked with green circles. The maximal-risk group (dLR+LR)-10 achieved a specificity of 96% at a sensitivity of 69%, resulting in an accuracy of 77%, which was statistically non-inferior to the accuracy of LR-5 ($p$ = 0.02). This result suggests that dLI-RADS and LI-RADS possibly capture complementary information: agreement between the model and radiological assessment at the highest risk level identifies a subset of lesions with very high HCC likelihood. However, this exploratory finding requires independent external validation.

## Feature importance

| HCC- DL | APHE-HF | NPW-HF | Size | NPW-DL | APHE-DL | EC-HF | EC-DL | AUC |
|---|---|---|---|---|---|---|---|---|
| ✓ | ✓ | ✓ | ✓ | ✓ | ✓ | ✓ | ✓ | **75.6** |
| ✓ | ✓ | ✓ | ✓ | ✓ | ✓ | ✓ | | **75.2** |
| ✓ | ✓ | ✓ | ✓ | ✓ | ✓ | | | 74.9 |
| ✓ | ✓ | ✓ | ✓ | ✓ | | | | 75.1 |
| ✓ | ✓ | | ✓ | ✓ | | | | 74.5 |
| ✓ | ✓ | ✓ | | | | | | 71.4 |
| ✓ | ✓ | | | | | | | 70.8 |
| ✓ | | | | | | | | 70.6 |

**Table 3**. Area under the receiver operating characteristic curve (AUC) diagnostic performance in ablation studies were the total number of dLI-RADS intermediate features is reduced from 8 to 1. For each configuration, the best-performing combination of features was selected using the train set and tested on the internal evaluation set. APHE: arterial phase hyper-enhancement, EC: enhancing capsule, NPW: non-peripheral washout, HF: handcrafted feature, DL: deep learning.

Ablation experiments assessed the contribution of the intermediate features (Table 3). The complete dLI-RADS model, using all eight intermediate features, achieved the highest area under the receiver operating characteristic curve on the internal evaluation set (75.6%). The HCC-DL feature was consistently selected across reduced configurations and achieved an AUC of 70.6% when used alone. Adding handcrafted and deep LI-RADS-inspired features improved performance, supporting the relevance of combining data-driven representations with clinically structured descriptors.

# Discussion

In this study, we proposed and evaluated dLI-RADS, a deep learning method for lesion-level HCC risk quantification that mimics the two-step paradigm of LI-RADS: imaging features are first estimated and then combined in an interpretable way to classify liver lesions into risk groups. dLI-RADS was non- inferior to LI-RADS in terms of HCC diagnostic accuracy on the internal evaluation set (80% vs. 74%, $p = 0.04$), and in terms of sensitivity on the external evaluation set (93% vs. 78%, $p < 0.001$). On this independent external evaluation set, which differed in patient demographics and disease etiologies, dLI-RADS maintained good standalone diagnostic performance, with an accuracy of 86%. However, because LI-RADS features were unavailable for non-HCC lesions, only sensitivity could be directly compared between dLI-RADS and LI-RADS, precluding a complete external comparison of their diagnostic performance. When combined, dLI-RADS and LI-RADS defined a maximal-risk category that achieved a very high specificity (96% at 69% sensitivity), suggesting possible complementarity between dLI-RADS and radiological LI-RADS assessment. However, this exploratory analysis was performed only on the internal evaluation set and requires confirmation in an independent external cohort.

A strength of our study is the comparison of a deep learning method against radiological interpretation using robust reference standards. In the main cohort, all nodules were analyzed histologically, and a composite reference standard was used to mitigate the limitations of negative biopsies. In the secondary cohort, all HCC nodules had histological confirmation, and non-HCC lesions were followed longitudinally for at least six months. By contrast, some prior studies such as [14] reported only the standalone performance of their proposed deep learning models, with limited or no comparison to radiologic readings. Our findings are consistent with prior work [12,14,20] in demonstrating that deep learning methods can match the diagnostic performance of radiologists but differ in that we formally tested for non-inferiority. This study specifically focused on small liver nodules (≤40 mm), with mean diameters of 18 mm and 22 mm in the internal and external evaluation sets respectively. We believe that this focus is especially relevant, as (i) radiologists are generally confident diagnosing lesions larger than 40 mm unaided, and (ii) curative options are limited and less effective for large-size HCC nodules. As exemplified by the accuracy achieved by dLI-RADS on our internal and external evaluation sets, 80% and 86% respectively, we also believe that lesion size is a major determinant of performance for automated diagnostic systems. To the best of our knowledge, our study is the first to report performance results for AI-based HCC diagnosis on an evaluation set of liver lesions with an average diameter inferior to 20 mm – referring here to our internal evaluation set. By contrast, Ho Yu et al. [14] reported a mean HCC size superior to 40 mm, and Ying et al. [20] reported mean sizes superior to 60 mm. By emulating LI-RADS' two-step paradigm, dLI-RADS adds a level of explainability to traditional deep learning-based classification. Among the 8 proposed intermediate features computed with dedicated neural networks or handcrafted radiomics, 6 can be interpreted as either APHE, EC, or NPW probabilities. This allows radiologists to compare their own assessments of LI-RADS major features with those inferred by the model. In contrast to our work, most reviewed deep learning models exclusively rely on end-to-end paradigms that offer little to no insight into the decision-making process [13,18, 20]. A notable exception is Ho Yu et al. [14], where a Grad-CAM approach provides visual explainability.

Our study has limits. First, the interpretability of the intermediate dLI-RADS features is only partial, as the most beneficial one is computed with a standard black-box deep learning approach trained in an end-to-end fashion. A feature importance analysis revealed that this component played a major role in the final dLI-RADS risk score. We argue however that our method, which encapsulates this black-box model into a two-step pipeline, offers an original compromise between performance and explainability.

From a technical perspective, it is interesting to note that complementing this deep learning model with more interpretable LI-RADS-inspired features was beneficial in terms of diagnostic performance. Secondly, in this proof-of-concept study, dLI-RADS relied on manual segmentations of the liver lesions to make predictions, which would not be realistically practicable in clinical routine. Moreover, because repeated independent segmentations were not available, we could not quantify the effect of inter-reader segmentation variability on dLI-RADS predictions or directly compare this source of variability with the inter-observer variability of conventional LI-RADS assessment. This issue may be particularly relevant for small lesions and should be specifically investigated in future studies. The present study design allowed us to focus on diagnostic endpoints independently of potential inaccuracies arising from automatic lesion segmentation. Future work should evaluate dLI-RADS as part of an end-to-end pipeline incorporating volumetric lesion segmentation. Thirdly, the number of patients in our evaluation cohorts was limited. Due to differing annotation and reference standard protocols, the two cohorts leveraged in this retrospective study could not be pooled together. Only the sensitivity of LI-RADS could be evaluated on the secondary cohort because LI-RADS major features were unavailable for non-HCC lesions. Consequently, although the full diagnostic performance of dLI-RADS could be assessed externally, the external cohort did not allow a complete comparison of dLI-RADS and LI-RADS in terms of specificity or accuracy. We underline however that validation data were collected from a total of 17 centers, in both Europe and Asia.

To conclude, dLI-RADS achieved promising performance for the AI-based lesion-level diagnosis of early HCC in CT and was non-inferior to LI-RADS on challenging evaluation sets of small-size liver lesions. When combined, dLI-RADS and LI-RADS defined a maximal-risk category that achieved a very high specificity, highlighting possible complementarity between model-based and radiological assessments, which requires confirmation through independent external validation. Prospective studies are warranted to assess the utility of dLI-RADS in clinical workflows and its ability to support decision-making in real-world settings.

## Data availability

The data underlying this article will be shared on reasonable request to the corresponding author.

## Code availability

The source code and pretrained model weights used in this study are publicly available on GitHub at https://github.com/esarf/dlirads and archived on Zenodo at https://doi.org/10.5281/zenodo.22667588 [28].

## Acknowledgements

Funding was provided by Région Île-de-France (ChoTherIA project) and ANRT (CIFRE #2021/1735), with additional high-performance computational resources from IDRIS under GENCI allocations 2023-A0150314655 and 2023-AD011014090.

## Author contributions

Conceptualization was led by Alexandre Bône, Pietro Gori, François Nicolas and Christophe Aubé, with support from Emma Sarfati, Marc-Michel Rohé, Jeong-Min Lee, Jeong Hee Yoon and Maxime Ronot.

Data curation was led by Emma Sarfati and Maxime Ronot, with support from Alexandre Bône, Jeong-Min Lee and Jeong Hee Yoon.

Project administration was led by Alexandre Bône, Marc-Michel Rohé, François Nicolas and Maxime Ronot.

Formal analysis was led by Emma Sarfati, with support from Alexandre Bône.

Resources were led by Pietro Gori, Marc-Michel Rohé, François Nicolas, Jeong-Min Lee and Christophe Aubé, with support from Maxime Ronot, Isabelle Bloch.

Investigation was performed by Emma Sarfati.

Methodology was led by Alexandre Bône, Pietro Gori, Marc-Michel Rohé, Isabelle Bloch and Christophe Aubé, with support from Emma Sarfati, Jeong Hee Yoon and Maxime Ronot.

Software was developed by Emma Sarfati and Alexandre Bône.

Supervision was led by Pietro Gori, François Nicolas and Isabelle Bloch.

Validation was led by Pietro Gori, Sisi Yang, Marc-Michel Rohé, Jeong-Min Lee, Maxime Ronot, Isabelle Bloch and Christophe Aubé, with support from Jeong Hee Yoon.

The original draft was written by Emma Sarfati and Alexandre Bône.

Writing - review & editing were led by Alexandre Bône, Emma Sarfati, Pietro Gori, Sisi Yang, Marc-Michel Rohé, Jeong-Min Lee, Jeong Hee Yoon, Isabelle Bloch and Christophe Aubé, with support from François Nicolas and Maxime Ronot.

## Competing interests

Emma Sarfati, Alexandre Bône, Sisi Yang, Marc-Michel Rohé and François Nicolas are employed by Guerbet. The remaining authors declare no competing interests.

# Figure legends

**Figure 1**. Overview of the dLI-RADS method. Tri-phasic computed tomography patches and lesion masks are used to compute handcrafted descriptors and deep learning-derived probabilities related to LI-RADS major features and HCC presence. These intermediate features are combined with logistic regression to predict lesion-level HCC risk.

**Figure 2**. Flowchart of patient inclusion. The diagram shows the inclusion and exclusion steps for the main and external cohorts, including exclusions for incomplete multiphase computed tomography, lesion size greater than 40 mm, missing definitive diagnosis, and LI-RADS inapplicability.

**Figure 3**. Examples of liver lesions analyzed by dLI-RADS and LI-RADS. Representative computed tomography patches are shown with corresponding radar plots summarizing model-derived and radiological features. The figure illustrates concordant and discordant assessments between dLI-RADS and LI-RADS.

**Figure 4**. Receiver operating characteristic curves for dLI-RADS. Curves are shown for the internal **(a)** and external **(b)** evaluation sets. Operating points corresponding to dLR-5 and dLR-4,5 are shown together with the corresponding LI-RADS operating points when available.

**Figure 5**. Combined dLI-RADS and LI-RADS assessment. Compound risk groups were defined by summing dLI-RADS and LI-RADS categories on the internal evaluation set. The maximal combined category reached high specificity, suggesting possible complementarity between model-based and radiological assessment.

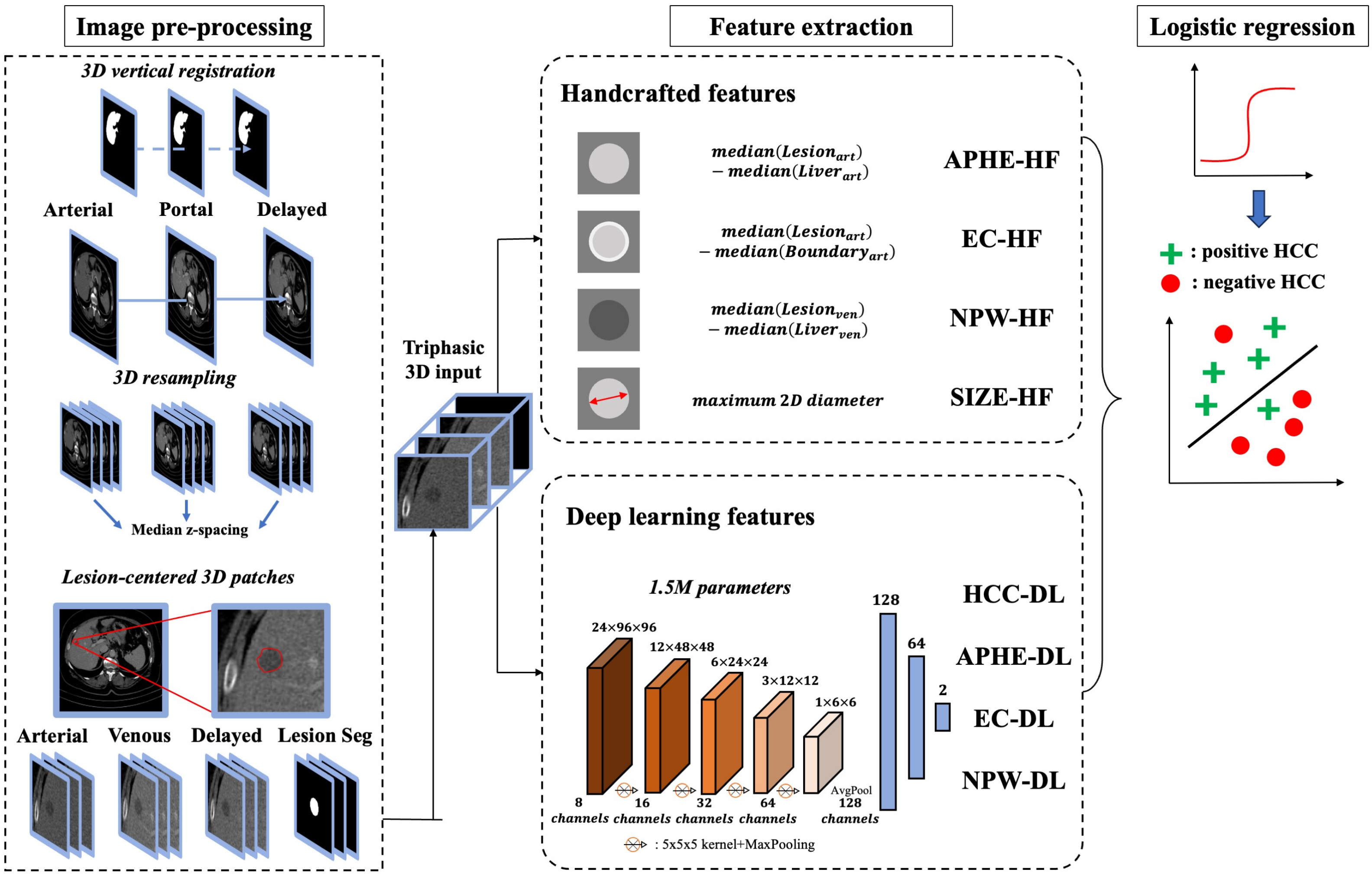
Image pre-processing
3D vertical registration
Arterial
Portal
Delayed
3D resampling
Median z-spacing
Lesion-centered 3D patches
Arterial
Venous
Delayed
Lesion Seg
Triphasic
3D input
Feature extraction
Handcrafted features
median(Lesion_art) − median(Liver_art)
APHE-HF
median(Lesion_art) − median(Boundary_art)
EC-HF
median(Lesion_ven) − median(Liver_ven)
NPW-HF
maximum 2D diameter
SIZE-HF
Deep learning features
1.5M parameters
24×96×96
12×48×48
6×24×24
3×12×12
1×6×6
8 channels
16 channels
32 channels
64 channels
AvgPool 128 channels
128
64
2
: 5x5x5 kernel+MaxPooling
HCC-DL
APHE-DL
EC-DL
NPW-DL
Logistic regression
: positive HCC
: negative HCC

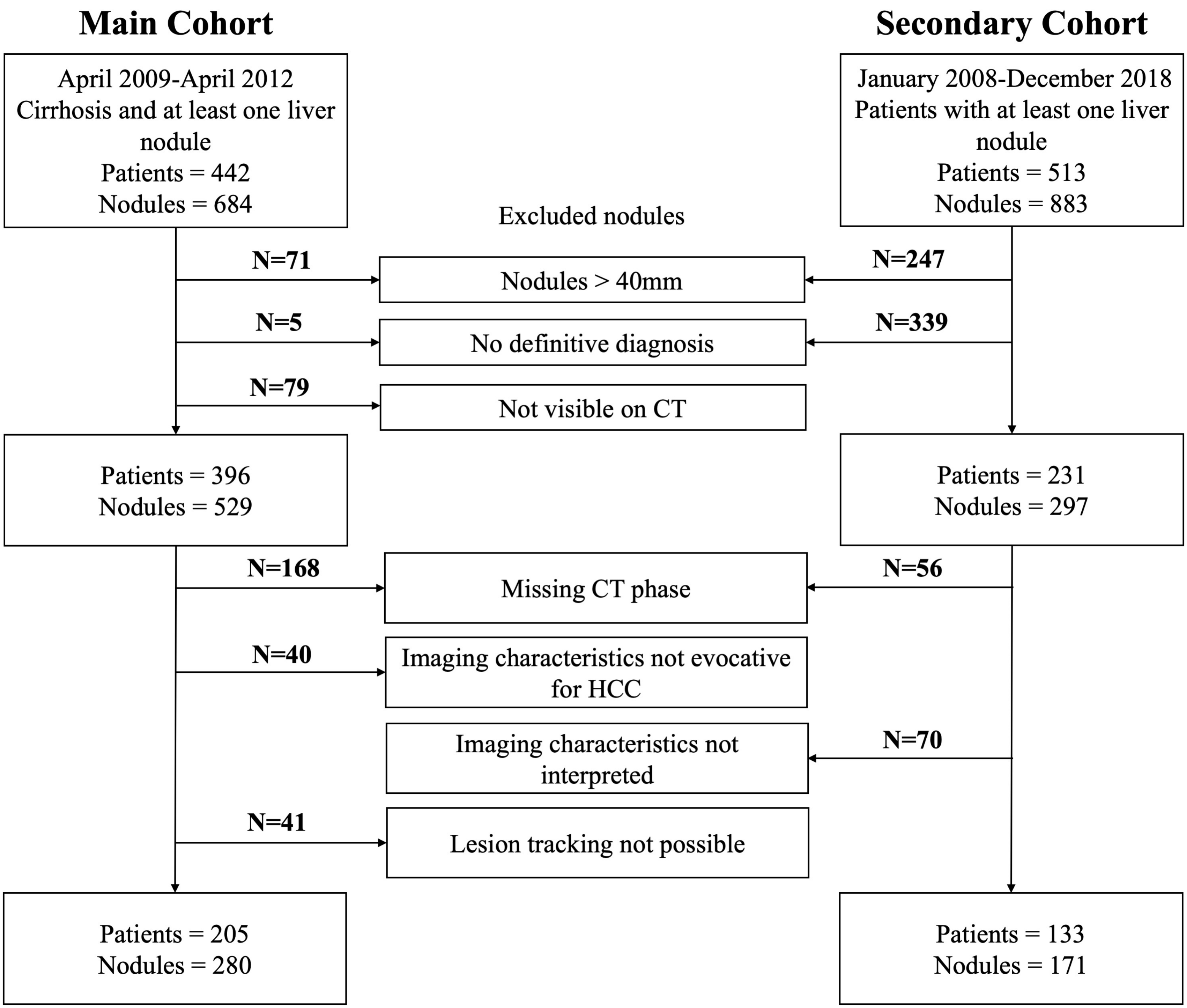
Main Cohort
Secondary Cohort
April 2009-April 2012
Cirrhosis and at least one liver nodule
Patients = 442
Nodules = 684
January 2008-December 2018
Patients with at least one liver nodule
Patients = 513
Nodules = 883
Excluded nodules
N=71
Nodules > 40mm
N=247
N=5
No definitive diagnosis
N=339
N=79
Not visible on CT
Patients = 396
Nodules = 529
Patients = 231
Nodules = 297
N=168
Missing CT phase
N=56
N=40
Imaging characteristics not evocative for HCC
Imaging characteristics not interpreted
N=70
N=41
Lesion tracking not possible
Patients = 205
Nodules = 280
Patients = 133
Nodules = 171

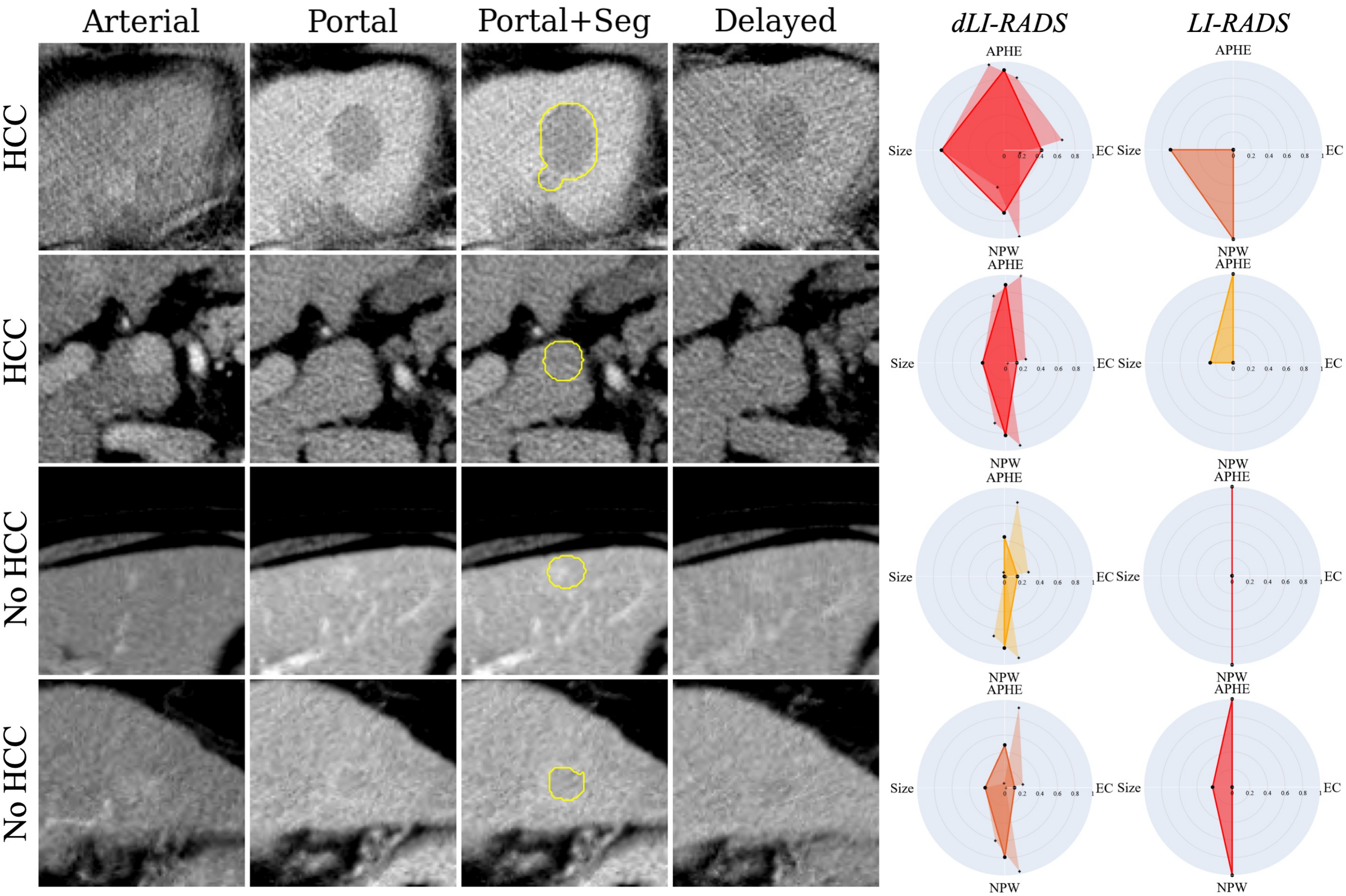
Arterial
Portal
Portal+Seg
Delayed
dLI-RADS
LI-RADS
HCC
HCC
No HCC
No HCC
APHE
Size
EC
NPW

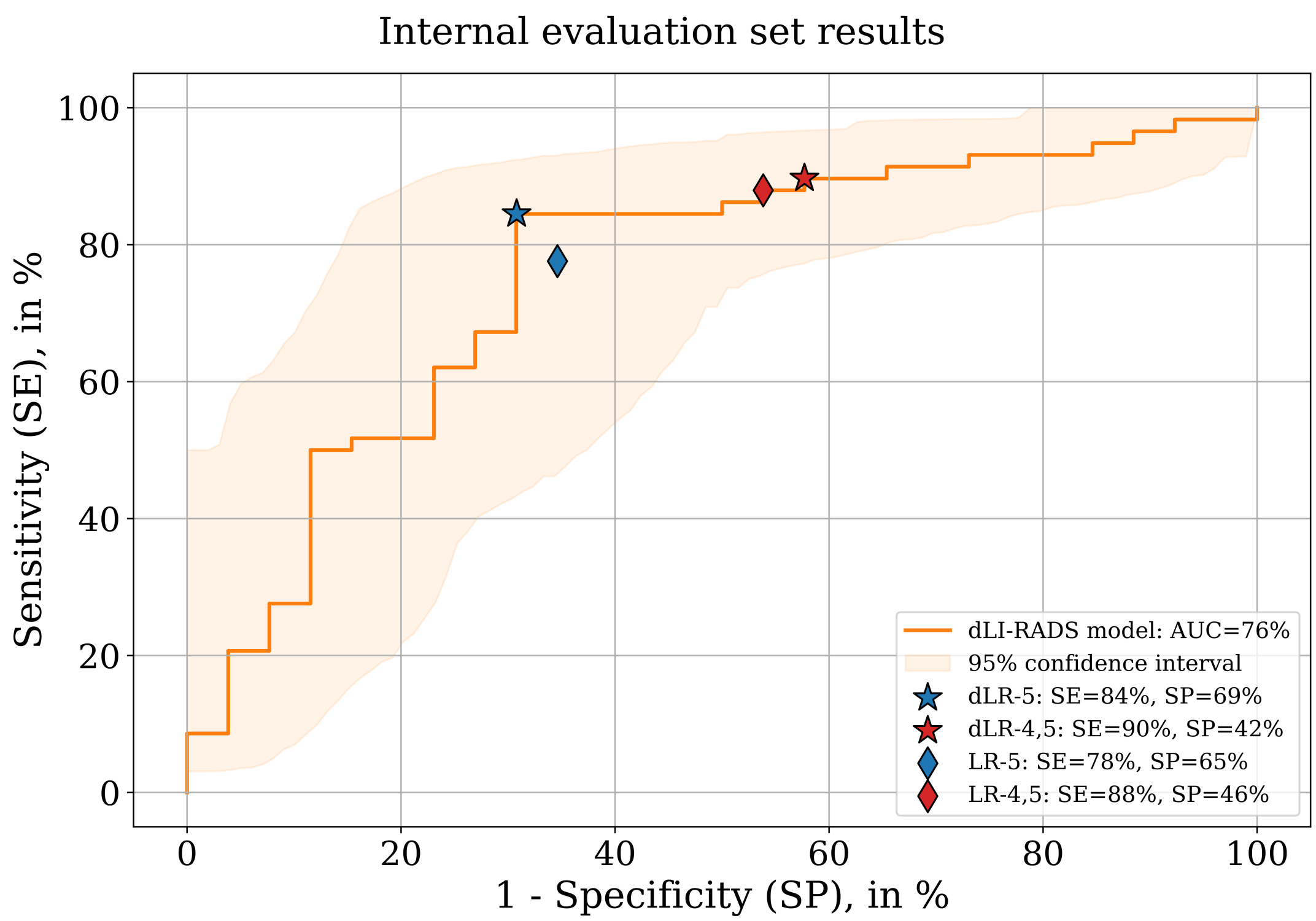
Internal evaluation set results
Sensitivity (SE), in %
1 - Specificity (SP), in %
0
20
40
60
80
100
dLI-RADS model: AUC=76%
95% confidence interval
dLR-5: SE=84%, SP=69%
dLR-4,5: SE=90%, SP=42%
LR-5: SE=78%, SP=65%
LR-4,5: SE=88%, SP=46%

**(a)**

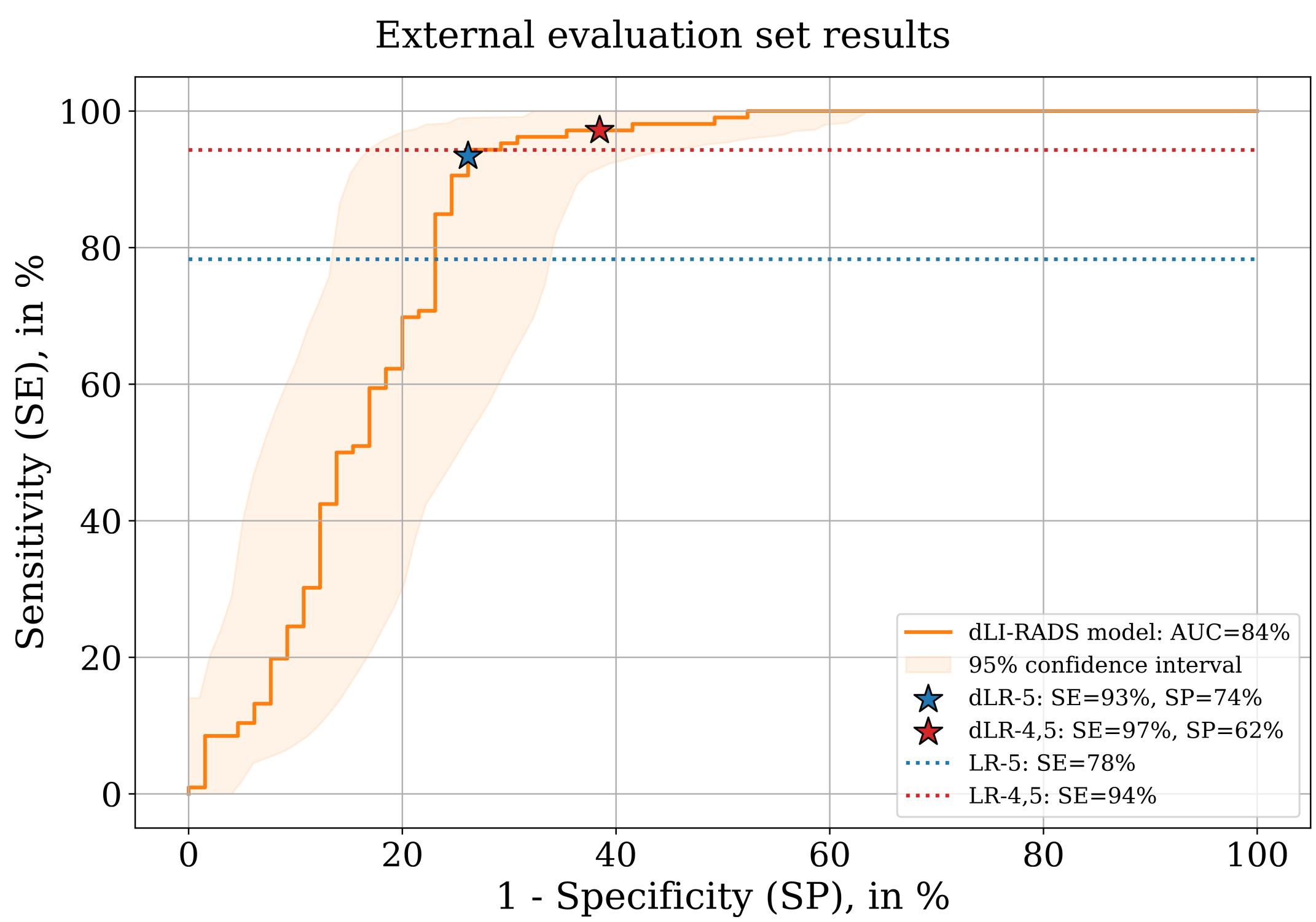
External evaluation set results
Sensitivity (SE), in %
1 - Specificity (SP), in %
0
20
40
60
80
100
dLI-RADS model: AUC=84%
95% confidence interval
dLR-5: SE=93%, SP=74%
dLR-4,5: SE=97%, SP=62%
LR-5: SE=78%
LR-4,5: SE=94%

**(b)**

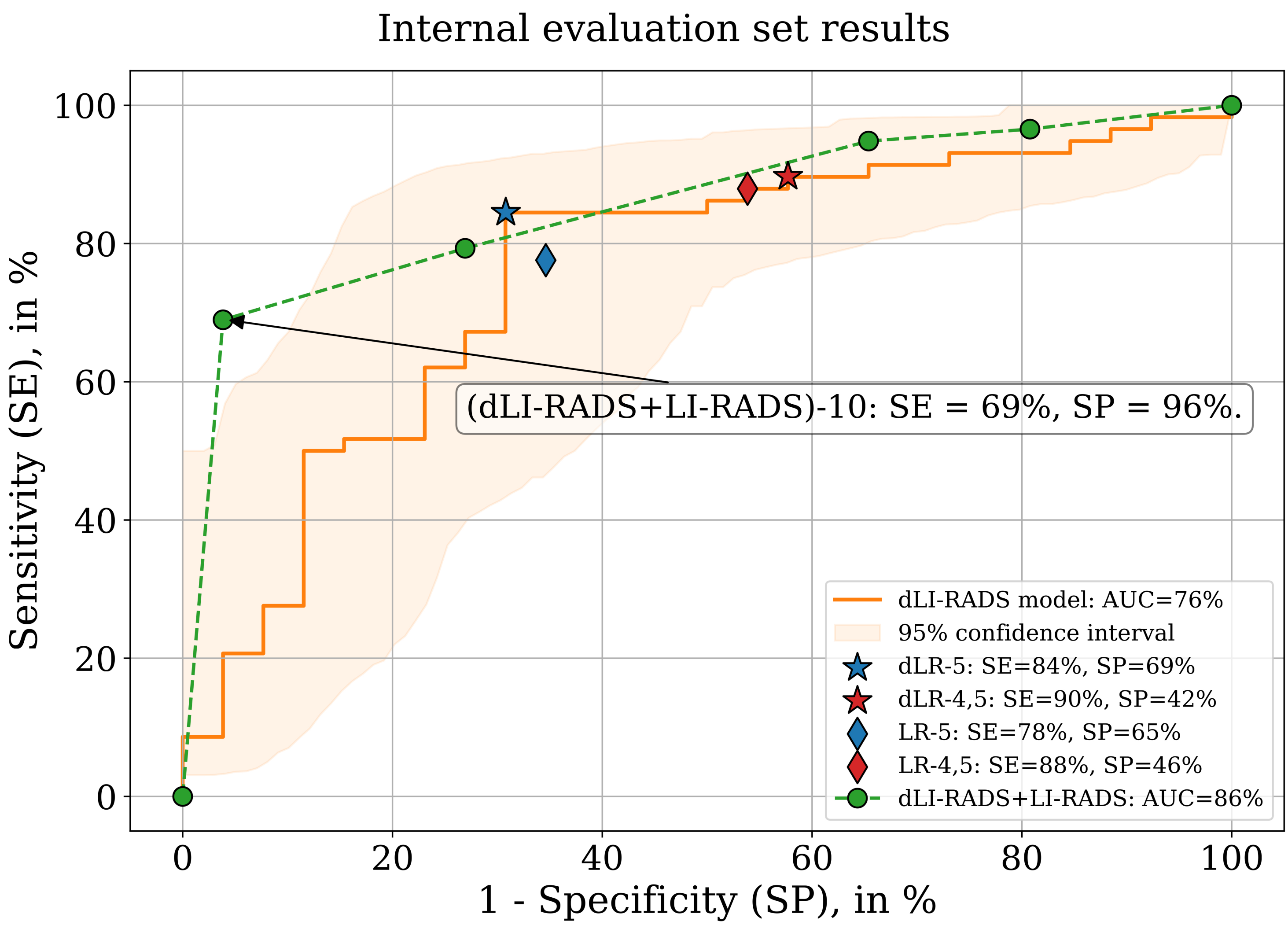
Internal evaluation set results
Sensitivity (SE), in %
1 - Specificity (SP), in %
0
20
40
60
80
100
(dLI-RADS+LI-RADS)-10: SE = 69%, SP = 96%.
dLI-RADS model: AUC=76%
95% confidence interval
dLR-5: SE=84%, SP=69%
dLR-4,5: SE=90%, SP=42%
LR-5: SE=78%, SP=65%
LR-4,5: SE=88%, SP=46%
dLI-RADS+LI-RADS: AUC=86%

# Tables

| | Training set (*N=143*) | Internal evaluation set (*N=62*) | External evaluation set (*N=133*) |
|---|---|---|---|
| Age, years, range | 61.2 (40.2,82.2) | 61.3 (41.2,81.1) | 58.9 (32.1,88.5) |
| Sex (Male / Female / Unknown) | 87.4% / 12.6% / 0% (125 / 18 / 0) | 79.0% / 21.0% / 0.0% (49 / 13 / 0) | 77.4% / 22.6% / 0.0% (103 / 30 / 0) |
| Number of centers | 16 | 16 | 1 |
| Liver cirrhosis (Yes / No / Unknown) | 100.0% / 0.0% / 0.0% (143 / 0 / 0) | 100.0% / 0.0% / 0.0% (62 / 0 / 0) | 88.7% / 3.8% / 7.5% (118 / 5 / 10) |
| Risk factor | | | |
| Alcohol (Yes / No / Unknown) | 57.3% / 41.3% / 1.4% (82 / 59 / 2) | 42.0% / 58.0% / 0% (26 / 36 / 0) | 8.3% / 91.7% / 0.0% (11 / 122 / 0) |
| Metabolic dysfunction-associated steatohepatitis (Yes / No / Unknown) | 18.9% / 79.7% / 1.4% (27 / 114 / 2) | 13.0% / 87.0% / 0.0% (8 / 54 / 0) | 67.7% / 21.8% / 10.5% (90 / 29 / 14) |
| Hepatitis B (Yes / No / Unknown) | 5.6% / 93% / 1.4% (8 / 133 / 2) | 6.5% / 93.5% / 0.0% (4 / 58 / 0) | 76.7% / 13.5% / 9.8% (102 / 18 / 13) |
| Hepatitis C (Yes / No / Unknown) | 21% / 77.6% / 1.4% (30 / 111 / 2) | 35.5% / 64.5% / 0.0% (22 / 40 / 0) | 8.3% / 82% / 9.7% (11 / 109 / 13) |
| Multi-factor (Yes / No / Unknown) | 10.5% / 88.1% / 1.4% (15 / 126 / 2) | 11.3% / 88.7% / 0.0% (7 / 55 / 0) | Missing values |
| Child-Pugh (A / B / C / Unknown) | 69.9% / 20.9% / 6.1% / 3.1% (137 / 41 / 12 / 6) | 69% / 19% / 7.1% / 4.9% (58 / 16 / 6 / 4) | Missing values |
| Lesions per patient (Mean ± Std) | 1.4 ± 0.6 | 1.4 ± 0.6 | 1.3 ± 1.0 |
| Lesion size (Mean ± Std) (mm) | 17.7 ± 6.0 | 18.3 ± 5.8 | 21.8 ± 9.5 |
| Lesion volume (Mean ± Std) ($cm^3$) | 4.5 ± 4.0 | 6.1 ± 6.0 | 4.9 ± 5.1 |
| Lesion count (N / %)<br><br>Lesion size (Mean ± Std) | | | |
| Hepatocellular carcinoma (HCC) | 130 / 66.3% 18.8 ± 5.8 | 58 / 69.0% 19.7 ± 5.5 | 106 / 62.0% 26.0 ± 6.8 |
| Simple cyst or lipoma | 7 / 3.6% 14.1 ± 3.6 | 4 / 4.8% 18.5 ± 7.4 | 39 / 22.8% 9.7 ± 2.7 |
| Metastasis | 14 / 7.1% 18.6 ± 8.4 | 5 / 6% 13.2 ± 2.1 | 5 / 2.9% 18.4 ± 8.4 |
| Hemangioma | 30 / 15.3% 14.8 ± 5.1 | 14 / 16.7% 15.1 ± 5.5 | 5 / 2.9% 15.9 ± 6.2 |
| Cholangiocarcinoma | / | / | 14 / 8.2% 27.1 ± 10.8 |
| Focal nodular hyperplasia | 1 / 0.5% 21.0 ± 0.0 | / | 1 / 0.6% 20.3 ± 0.0 |
| Perfusion defect | / | / | 1 / 0.6% 12.1 ± 0.0 |
| Other | 14 / 7.1% 13.7 ± 3.9 | 3 / 3.6% 16.5 ± 5.0 | / |

**Table 1**. Clinical characteristics of patients in training, internal evaluation, and external evaluation sets.

| | Internal evaluation set (N=62) | | | | | | External evaluation set (N=133) | | | | | |
|---|---|---|---|---|---|---|---|---|---|---|---|---|
| | dLI-RADS | | LI-RADS | | Non-inferiority p-values | | dLI-RADS | | LI-RADS | | Non-inferiority p-values | |
| | dLR-4,5 | dLR-5 | LR-4,5 | LR-5 | dLR-4,5 vs. LR4,5 | dLR-5 vs LR-5 | dLR-4,5 | dLR-5 | LR-4,5 | LR-5 | dLR-4,5 vs. LR4,5 | dLR-5 vs LR-5 |
| SE | 90<br>(81,97)<br>[52/58] | 84<br>(75,93)<br>[49/58] | 88<br>(79,96)<br>[51/58] | 78<br>(67,88)<br>[45/58] | 0.10 | **0.03** | 97<br>(94,100)<br>[103/106] | 93<br>(88,98)<br>[99/106] | 94<br>(90,98)<br>[100/106] | 78<br>(70,86)<br>[83/106] | **0.001** | **<0.001** |
| SP | 42<br>(23,62)<br>[11/26] | 69<br>(50,86)<br>[18/26] | 46<br>(27,65)<br>[12/26] | 65<br>(46,83)<br>[17/26] | 0.44 | 0.26 | 62<br>(49,73)<br>[40/65] | 74<br>(63,84)<br>[48/65] | n/a | | | |
| ACC | 75<br>(66,83)<br>[63/84] | 80<br>(70,88)<br>[67/84] | 75<br>(66,83)<br>[63/84] | 74<br>(64,83)<br>[62/84] | 0.16 | **0.04** | 84<br>(78,89)<br>[143/171] | 86<br>(81,91)<br>[147/171] | | | | |
| AUC | 76 (63,87) | | n/a | | | | 84 (76,91) | | | | | |

**Table 2**. Diagnostic performance of dLI-RADS and LI-RADS for HCC diagnosis on internal evaluation and external evaluation sets. Diagnostic performance of LI-RADS (LR) and the proposed dLI-RADS (dLR) are assessed on two risk groups: high-risk (dLR-4,5, LR-4,5) and maximal-risk (dLR-5, LR-5). Performance metrics are sensitivity (SE), specificity (SP), accuracy (ACC), and area under the receiver operating characteristic curve (AUC). All metrics are expressed in percentage points. Between parentheses are given 95% confidence intervals, and between brackets the integer numerator and denominator. Non-inferiority *p*-values of dLI-RADS vs. LI-RADS computed with 10,000-repetition bootstrap testing are also given in dedicated columns. Non-inferiority margin and significance level were both 5%. Bold and underlined *p*-values indicate statistical significance. Reported *p*-values correspond exclusively to non-inferiority testing and should not be interpreted as tests of superiority.

| HCC- DL | APHE-HF | NPW-HF | Size | NPW-DL | APHE-DL | EC-HF | EC-DL | AUC |
|---|---|---|---|---|---|---|---|---|
| ✓ | ✓ | ✓ | ✓ | ✓ | ✓ | ✓ | ✓ | **75.6** |
| ✓ | ✓ | ✓ | ✓ | ✓ | ✓ | ✓ |  | **75.2** |
| ✓ | ✓ | ✓ | ✓ | ✓ | ✓ |  |  | 74.9 |
| ✓ | ✓ | ✓ | ✓ | ✓ |  |  |  | 75.1 |
| ✓ | ✓ |  | ✓ | ✓ |  |  |  | 74.5 |
| ✓ | ✓ | ✓ |  |  |  |  |  | 71.4 |
| ✓ | ✓ |  |  |  |  |  |  | 70.8 |
| ✓ |  |  |  |  |  |  |  | 70.6 |

**Table 3**. Area under the receiver operating characteristic curve (AUC) diagnostic performance in ablation studies were the total number of dLI-RADS intermediate features is reduced from 8 to 1. For each configuration, the best-performing combination of features was selected using the train set and tested on the internal evaluation set. APHE: arterial phase hyper-enhancement, EC: enhancing capsule, NPW: non-peripheral washout, HF: handcrafted feature, DL: deep learning.